\documentclass[journal,final]{IEEEtran}
\ifCLASSINFOpdf
\else
\fi
\usepackage{epsfig}
\usepackage{graphicx}
\usepackage{amsmath}
\usepackage{amssymb}
\usepackage{textcomp}
\usepackage{mathrsfs}
\usepackage{booktabs}
\usepackage{diagbox}
\usepackage{xcolor}
\usepackage{multirow}
\usepackage{subfigure}
\usepackage{footnote}
\usepackage{booktabs}
\usepackage{threeparttable}
\usepackage{ragged2e}
\usepackage{textcase}

\usepackage[pagebackref=false,breaklinks=true,letterpaper=true,colorlinks,bookmarks=false,linkcolor=red,citecolor=blue]{hyperref}

\begin{document}
	%
	\title{Semantic Change Detection with \\ Asymmetric Siamese Networks}
	%
	%
	%
	
	\author{Kunping~Yang,
		\and Gui-Song~Xia,~\IEEEmembership{Senior Member,~IEEE,} 
		\and Zicheng~Liu,
		\and Bo Du,~\IEEEmembership{Senior Member,~IEEE,} \\
		\and Wen Yang,~\IEEEmembership{Senior Member,~IEEE,}  
		\and Marcello Pelillo,~\IEEEmembership{Fellow,~IEEE,} 
		\and Liangpei Zhang,~\IEEEmembership{Fellow,~IEEE} 
		\IEEEcompsocitemizethanks{
			\IEEEcompsocthanksitem The study of this paper is funded by the National Natural Science Foundation of China (NSFC) under grant contracts No.61922065, No.61771350 and No.41820104006. It is also partially funded by the Major Projects of Technological Innovation in Hubei Province (2019AEA170). 
			\IEEEcompsocthanksitem K. Yang, G.-S. Xia, Z. Liu, B. Du and L. Zhang are with the State Key Lab. of LIESMARS and the 
			School of Computer Science, Wuhan University, Wuhan, 430072, China.  Email: \{{\em kunpingyang, guisong.xia, zicheng.liu, dubo, zlp62}\}@whu.edu.cn.
			\IEEEcompsocthanksitem W. Yang is with the School of Electronic Information, Wuhan University, Wuhan, 430072, China.  Email: yangwen@whu.edu.cn.
			\IEEEcompsocthanksitem M. Pelillo is with DAIS, University of Venice, 30172, Italy.  
			Email: pelillo@unive.it.
			\IEEEcompsocthanksitem Corresponding author: Gui-Song Xia (guisong.xia@whu.edu.cn).
		}
	}

	\maketitle
	
	\begin{abstract}
		\justifying
		Given two multi-temporal aerial images, semantic change detection aims to locate the land-cover variations and identify their change types with pixel-wise boundaries. This problem is vital in many earth vision related tasks, such as precise urban planning and natural resource management.
		Existing state-of-the-art algorithms mainly identify the changed pixels by applying homogeneous operations on each input image and comparing the extracted features. However, in changed regions, totally different land-cover distributions often require heterogeneous features extraction procedures \emph{w.r.t} each input. In this paper, we present an {\em asymmetric siamese network} (ASN) to locate and identify semantic changes through feature pairs obtained from modules of widely different structures, which involve areas of various sizes and apply different quantities of parameters to factor in the discrepancy across different land-cover distributions. To better train and evaluate our model, we create a large-scale well-annotated {\em {\color{black}SE}mantic {\color{black}C}hange detecti{\color{black}ON} {\color{black}D}ataset} (SECOND), while an \emph{Adaptive Threshold Learning} (ATL) module and a {\em Separated Kappa} (SeK) coefficient are proposed to alleviate the influences of label imbalance in model training and evaluation. The experimental results demonstrate that the proposed model can stably outperform the state-of-the-art algorithms with different encoder backbones.
	\end{abstract}
	

	%
	\IEEEpeerreviewmaketitle

	\section{Introduction}\label{sec:introduction}
	%
	%
	%
	%
	\IEEEPARstart{C}{hange} detection in multi-temporal aerial images~\cite{A-Contrario,Statistic_PAMI,HRSCD,Systematic_Survey_TIP, Deep_Active_Learning}, which aims to locate and analyze the regions of land-cover variations on the earth surface, is a crucial image interpretation task related to many applications such as, precise urban planning~\cite{PollockPOLICY} and natural resource management~\cite{AlmutClimate,TownshendGlobal,BelwardHigh}. Given a pair of multi-temporal images, most existing methods focus on detecting the locations of changed pixels between the input images, namely binary change detection (BCD) {\em e.g.}~\cite{A-Contrario, Statistic_PAMI, ICCV_98_CD, Registration_PAMI}. However, since it overlooks pixels' categories, BCD often fails to depict the semantic change information that are highly demanded in subsequent applications. Hence, developing methods that can simultaneously extract changed regions and identify their land-cover classes in multi-temporal images, {\em i.e.} semantic change detection (SCD), has become an active research topic in recent years~\cite{HRSCD,Mou_TGRS}.
	
	An intuitive solution to achieve SCD is to first partition the input images into semantic regions 
	and then compare the segmentation results for identifying change types. 
	However, this direct solution makes the underlying assumption that semantic categories in multi-temporal images are independent and is typically troubled by two problematic aspects: 1) the changed regions of the same semantic category cannot be distinguished; 2) the intrinsic correlation across categories will be overlooked.
	
	\begin{figure}[!t]
		\centering
		{\includegraphics[width=1.01\linewidth]{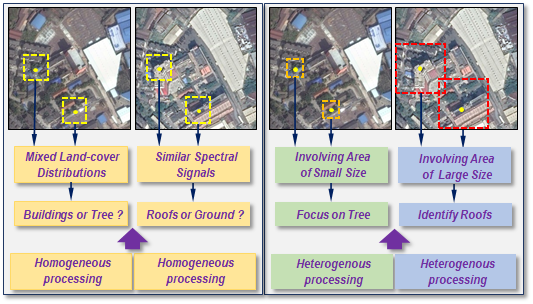}}
		\vspace{-5mm}
		\caption{In aerial images, land-cover objects appearing at different geometrical structures and mixed distributions across multi-temporal images, which we call asymmetric changes, make it difficult to locate and analyze land-cover variations through existing methods with homogeneous image processings \emph{w.r.t} each input.
			In contrast with existing methods, we are motivated to design some heterogenous image processings, which we call locally asymmetric, to factor in the discrepancy across different land-cover distributions and provide extra information for SCD problem.}
		\label{fig:Instance}
		\vspace{-2mm}
	\end{figure} 	
	In order to take advantage of categorical correlation in multi-temporal images, many existing SCD methods rely on the architecture of siamese networks~\cite{siamese} and have reported promising results~\cite{HRSCD, Mou_TGRS}. 
	However, in contrast to the application scenarios in \cite{CD_BMCV, ChangeNet}, the involved siamese networks in SCD would face with difficulties in locating and identifying changed regions conforming to different mixed land-cover distributions in each multi-temporal image, which we call asymmetric changes.
	\begin{figure}[t!]
		\centering
		{\includegraphics[height=0.88\linewidth]{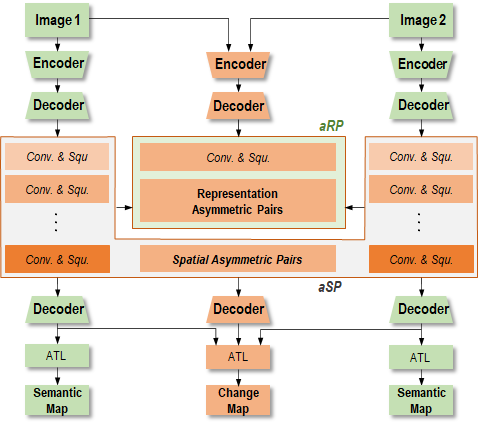}}
		\vspace{-3mm}
		\caption{{\em Asymmetric Siamese Network} (ASN) for SCD. ASN utilizes siamese encoders to map input multi-temporal images into feature space, while the siamese decoders are leveraged to obtain semantic maps. Similarly, encoder and decoders in change detection branch are designed to obtain change map. In contrast to traditional siamese network, ASN utilizes several convolutional sequences and squeeze gates in proposed aSP and aRP to obtain feature pairs deriving from widely different structures, which we call asymmetric feature pairs, to provide extra information. Furthermore, the designed ATL is exploited to adaptively revise the output deflections based on the combinations of raw model outputs through slight extra convolutional layers.}
		\vspace{-2mm}
		\label{fig:Process}
	\end{figure}
	For instance, focusing on the marked positions (yellow points) in Fig.\ref{fig:Instance}, to better identify regions with mixed distributions of \emph{tree} and \emph{buildings}, concentrating on small area (within orange boxes) could provide delicate details.
	But to alleviate the categorical ambiguity between \emph{impervious surface} and the roofs of \emph{buildings}, features involving more surroundings (within red boxes) are preferred.
	Furthermore, feature representation capabilities are often required to be adaptive.
	For example, pixels in one image belonging to objects with complex structures ({\em e.g. buildings} mixed with {\em tree}) often prefer modules with stronger representation capabilities, which however may make obvious flaws on those with simple structures ({\em e.g.} single {\em impervious surface}) in another image due to the over-fitting.
	In summary, mixed land-cover classes and various land-cover distributions often make object properties in each input image different, and cannot be modeled well by solely using symmetric architectures as in siamese networks. 
	Thus, it is of great importance to design a deep model that can better depict the semantic but asymmetric changes in images. 
	
	In this article, we address the asymmetric properties of SCD problem by exploiting siamese networks. As illustrated in Fig.~\ref{fig:Process}, we propose an {\em Asymmetric Siamese Network} (ASN) to extract changed pixels through two modules, {\em i.e.,} {\em{asymmetric Spatial Pyramid}} (aSP) and {\em{asymmetric Representation Pyramid}} (aRP).
	Leveraging designed convolution sequences of different structures, aSP and aRP obtain features through several siamese feature pyramids deriving from input images.
	Specifically, in aSP and aRP, we design weighted dense connected topological architectures, where different feature pairs across the obtained siamese feature pyramids deriving from each input are linked with various edges. Although the whole architecture is symmetric, most of these feature pairs are obtained by widely different structures, which we call asymmetric feature pairs and locally asymmetric structures. Dynamic branch weights further adjust the importance of each edge according to each input. Containing designated receptive fields and representation capabilities, these asymmetric feature pairs are able to focus on areas of various sizes and implicate different representation capabilities.
	As we shall see, compared with traditional siamese network, ASN can better depict asymmetric changes between mixed targets and illegible area.
	
	To better train and evaluate the proposed model, we create a well-annotated {\em SEmantic Change detectiON Dataset} (SECOND) to set up a new benchmark. Although existing SCD datasets contain abundant categorical information, they are often not big enough~\cite{Mou_TGRS}, which are inadequate to develop SCD algorithms with good generalization ability. Meanwhile, the annotations of some SCD datasets are unable to identify changes between the same land-cover class~\cite{HRSCD}. 
	The proposed SECOND is with $4662$ pairs of images in $30$ change types. Especially, we annotate the semantic categories and changed pixels separately in SECOND dataset, which makes changed regions between the same land-cover class available.
	
	Last but not least, model training and evaluation are always influenced by label imbalance due to the overwhelming categories, {\em e.g.} \emph{non-change} pixels. Specifically, severe label imbalance would make models tend to collapse. Thus, we propose an \emph{Adaptive Threshold Learning} (ATL) module to adaptively revise the deflections of semantic change outputs caused by label imbalance through learnable output adjustments, where the logical relationships between each semantic category can be explored.
	Moreover, existing metrics used to evaluate change detection algorithms, such as Overall Accuracy (OA) and Kappa coefficient ($\kappa$), are inherited from classification tasks, which would cause unreasonable scores due to the neglect of the dominant \emph{non-change} pixels. 
	Thus, we further present a {\em Separated Kappa} (SeK) coefficient as a modified evaluation measurement for semantic change detection task, which separates the \emph{non-change} class from other change types to reduce the effects of label imbalance. Compared with OA and $\kappa$, SeK is more in line with human scoring in SCD problem.
	
	\vspace{2mm}
	Our main contributions in this paper are threefold.
	\begin{itemize}
		\vspace{3mm}
		\item We propose an asymmetric siamese network, {\em i.e.} ASN, to factor in the discrepancy across different land-cover distributions in each multi-temporal image, which can alleviate the categorical ambiguity through extra information provided by heterogenous processings.
		\item We create a large-scale semantic change detection dataset, {\em i.e.} SECOND, to better train deep models and as a new benchmark for the SCD problem. This SECOND also enables us to distinguish changed regions between the same land-cover class.
		\item We design an \emph{Adaptive Threshold Learning} module and a {\em Separated Kappa}, {\em i.e.} ATL and SeK, to alleviate influences of label imbalance, which can adaptively revise the output deflections and fix unreasonable scores computed with traditional metrics, {\em e.g.} OA and $\kappa$, respectively.
		\vspace{3mm}
	\end{itemize}
	
	\section{Related Work}{\label{sec:related work}}
	\subsection{Location of Changed Regions}
	In order to locate changed regions in multi-temporal images, early works formulated the task as a BCD problem and relied on certain probability statistical  formulations~\cite{A-Contrario, Statistic_PAMI, Sequential_TIP, Multivariate_statistical_TIP, Statistical_TIP, Bivariate_Gamma_Distributions_TIP} or change vector analysis~\cite{Rayleigh_TIP, semiparametric_TIP, CVA, difference_image_analysis_tgrs, surveillance_TIP}. The statistical formulations and change vector analysis are able to explore the dissimilarity between features corresponding to changed and unchanged regions. Meanwhile, in order to embed pixel-wise correspondences, graphical model based algorithms are proposed \cite{MRF_CVPRW, Gibbs_MRF_TIP}, which can model the spatial regularity during the optimization process. Further to expand the model capacities, deep learning networks~\cite{Deep_Active_Learning, CD_FULLY_ICIP, MFCNET, SCCN, Procedural} are designed with elaborate structures to depict more diverse scenes by searching optimal parameters, where huge parameter space ensures stronger model capacities. Moreover, as illustrated in Sec.\ref{sec:introduction}, considering in the discrepancy across land-cover distributions in input images could provide extra information when depicting land-cover distributions in the cases shown in Fig.\ref{fig:Instance}.
	Thus, it is necessary to explore some asymmetric architectures in conventional deep networks. In this paper, we propose an ASN to leverage locally asymmetric structures, which would be able to detect changed regions through change detection branch more precisely compared with traditional siamese networks.

	\subsection{Identification of change types}
	Existing SCD algorithms mainly utilize two ways to identify the change type of each pixel. The first kind of models, {\em e.g.}~\cite{Mou_TGRS}, consider \emph{non-change} as a special change type, which extract unchanged regions and other change types simultaneously. The second kind of methods, {\em e.g.}~\cite{HRSCD, Innovative_Neural_Net_Method_tgrs}, separately extract unchanged regions and identify change types. However, these models often fail to consider issues such as multi-scale objects or scenes of various complexities in the interpretation process. As discussed in recent semantic segmentation algorithms~\cite{V3,V3+,HRNet}, parallel structures with diverse receptive fields significantly improve the model performances when dealing with multi-scale objects. In views of this insight, in SCD problem, we need to explore adaptive structures to obtain features suitable for different land-cover distributions corresponding to each input image. 
	Furthermore, dominant {\em non-change} pixels in SCD problem would make model tend to collapse during the training process~\cite{Long_Tail_CVPR}, which also leads us to investigate the revision process of raw model outputs. In proposed ASN, we design structures with various dilation rates and parameter quantities to adapt different land-cover distributions, while the proposed ATL learns to adaptively update semantic predictions in the light of raw model outputs.
	
	
	
	\subsection{Datasets and Evaluation Metrics}
	Benchmark datasets and evaluation metrics are two important aspects related to training the designed models and measuring the generalization capability of SCD algorithms. 
	
	Existing datasets, {\em e.g.}~\cite{Hermiston_City_area, quasi_urban_areas,OSCD,ACD}, are mainly created for the BCD problem. The lack of land-cover categorical information still limits their usage in SCD, although \cite{Deep_Active_Learning} has made an impressive contribution to alleviate data scarcity through active learning. A few benckmark datasets have been built for SCD~\cite{HRSCD,Mou_TGRS, Innovative_Neural_Net_Method_tgrs}. Among them, the SCD datasets used in~\cite{Mou_TGRS, Innovative_Neural_Net_Method_tgrs} are not big enough to sufficiently train and evaluate SCD algorithms. While, the dataset in~\cite{HRSCD} utilizes two independently annotated land-cover maps to represent the change types, which ignores changed regions between the same land-cover class. Thus, in our proposed SECOND dataset, we check multi-temporal images simultaneously to annotate land-cover classes and changed regions separately, which can distinguish changed regions between the same land-cover class.
	
	
	On the other hand, commonly used evaluation metrics for change detection tasks are mainly inherited from those used in classification problems, such as OA and $\kappa$ used in~\cite{HRSCD, Mou_TGRS}, which ignore the fact that unchanged regions are often of the overwhelming majority in change maps. As a consequence, models with dominant predictions of {\emph{non-change}} pixels would get unreasonable high scores in terms of OA and $\kappa$. 
	Thus, it is demanded to take into account the label imbalance for better evaluations.
	In this paper, we utilize mean Intersection Over Union (mIOU) \cite{Pascal} and design a new metric, {\em i.e.} SeK, to measure the SCD results, which can alleviate the label imbalance effects.
	
	
	\section{Methods}	
	\subsection{Problem Definition}
	Let $I_1, I_2:\Omega \rightarrow \mathbb{R}^{d}$ denote two multi-temporal images of $d \in \mathbb{N}_+$ channels, with $\Omega$ being the image grid $\{0,1,\ldots,H-1\} \times \{0,1,\ldots,W-1\}$. Given a set $L=\{y_1,\cdots, y_N\}$ of $N$ semantic categories, the SCD problem aims to find a mapping function $f_{I_1, I_2}: \Omega \mapsto L^2$ such that
	\begin{equation}\label{1}
	\forall \mathbf p \in \Omega, \ f_{I_1, I_2}(\mathbf p) = \left\{\begin{array}{ll}
	(0,\,0) & \text{if}\,~\mathcal{C}_{I_1, I_2}(\mathbf p) < \tau,\\
	(l_1,\,l_2), &
	\text{otherwise},
	\end{array}\right.
	\end{equation}
	where $\mathcal{C}_{I_1, I_2}(\mathbf p)$ measures the change probability of each pixel $\mathbf p \in \Omega$, $l_1, l_2 \in L$, and $(0,\, 0)$ indicates {\em non-change} class. $\tau$ is a scalar thresholding on $\mathcal{C}_{I_1, I_2}$. Thus, $f_{I_1, I_2}$ can locate changed regions and identify their categories simultaneously. 
	
	\subsection{An Intuitive Solution}\label{sec:Intuitive}
	An intuitive solution to obtain such a $f_{I_1, I_2}$ is to first partition the input images $I_1$ and $I_2$ into semantic regions, {\em e.g.} with semantic segmentation algorithms, and then compare them to locate and identify semantic changes.
	Specifically, for $t=1,2$, let $\mathcal{M}_t: \Omega \mapsto \mathbb{R}^N$ be the semantic probability map of $I_t$, {\em i.e.} the probability vector $\mathcal{M}_t(\mathbf p) \in \mathbb{R}^{N}$ indicates the possibility of pixel $\mathbf p \in \Omega$ belonging to each semantic category in $L$. 
	Denoting $l_t = \mathop{\arg\max_{l \in L}}{\mathcal{M}_t(\mathbf p)}$ as the semantic category of pixel $\mathbf p$ in $I_t$, we have
	\begin{equation}\label{2}
	\centering
	f_{I_1, I_2}(\mathbf p) = \left\{\begin{array}{ll}
	(0, \,0), & \text{if}~l_1 = l_2,\\
	(l_1, \, l_2), &
	\text{otherwise}.
	\end{array}\right.
	\end{equation}
	
	As $\max \mathcal{M}_1(\mathbf p) \cdot \max \mathcal{M}_2(\mathbf p) = \max (\mathcal{M}_1^{T}(\mathbf p) \times \mathcal{M}_2(\mathbf p))$, for Eq.~\eqref{2}, we have
	$$
	(l_1, l_2) = \arg\max_{(l_1, l_2) \in L^2}{(\mathcal{M}_1^{T}(\mathbf p) \times \mathcal{M}_2(\mathbf p))},
	$$
	where the operation $\times$ represents matrix product.	%
	However, due to the non-relevance between $\mathcal{M}_1$ and $\mathcal{M}_2$,
	the intuitive solution actually implies that {\em the semantic categories in each image are independent for every pixel}. However, this underlying assumption is quite different from the reality. The intrinsic correlation between each category is important and the model performance would be limited by overlooking the categorical correlation. Moreover, according to Eq.~\eqref{2}, this intuitive solution also can not identify changed regions between same land-cover classes due to the same output form with {\em non-change} pixels.
	\subsection{SCD with Conventional Siamese Networks}
	To take categorical correlation into account, the state-of-the-art algorithm, {\em i.e.} HRSCD.str4~\cite{HRSCD}, utilizes siamese semantic segmentation branches with an extra change detection branch to address the SCD problem. During the parameter optimization process, the siamese branches would influence each other through the gradient flows and skip connections.
	
	However, it is worth noticing that HRSCD.str4 relys on totally symmetric structures. More precisely, the response values in feature maps of change detection branch are related to the areas of same size and obtained by nearly the same mapping functions on multi-temporal images. Besides, the siamese decoders utilize single-line structure with skip connection, which are not fully aware of various object scale distributions \emph{w.r.t.} each input.
	As we shall see in Sec.\ref{sec:Experiments}, this symmetric architecture makes it difficult to locate and identify semantic changes related to widely different land-cover distributions across multi-temporal images in some asymmetric changes.
	
	
	\begin{figure}[t!]
		\centering
		{\includegraphics[width=1\linewidth]{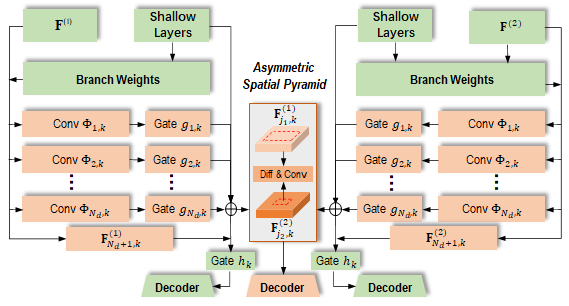}}	
		\vspace{-5mm}
		\caption{The proposed aSP module with length of 1. Index $k$ controls the channel numbers of layers in each convolution sequence, while $j_1$,$j_2$ indicate different receptive fields. Each squeeze gate consists of the concatenation operator, convolution layers and skip connections. aSP exploits asymmetric spatial feature pairs with diverse spatial information.}
		\vspace{-3mm}
		\label{fig:SAM}
	\end{figure}
	
	\subsection{Asymmetric Siamese Network for SCD}
	\subsubsection{Overall Architecture}
	In order to depict the aforementioned asymmetric changes that beyond the descriptive capability of conventional siamese networks, we propose to integrate features deriving from heterogenous processes that are adaptive to each input. Specifically, as illustrated in Fig.\ref{fig:Process}, the proposed ASN exploits siamese feature pairs \emph{w.r.t.} the input images, denoted as $I_1, I_2$, in semantic segmentation branches, based on which designed aSP and aRP generate feature pairs with various receptive fields and representation capabilities through convolution sequences with different dilation rates and parameter quantities. These feature pairs are further integrated into the change detection branch and semantic segmentation branches to obtain predictions of {\em non-change} pixels and semantic categories, which finally compose the semantic change detection results.

	\subsubsection{Asymmetric Spatial Pyramid (aSP)}
	In aSP, we design several parallel convolution sequences $\{\Phi_{j,k} \}_{1\le j\le N_d, \, 1 \le k \le N_r}$ of diverse structures, where each sequence is a convolutional operation set, denoted as $\Phi_{j,k}=\{\phi_{i,j,k}\}_{1 \le i \le N_c}$. Specifically, $N_d, N_c$ and $N_r$ are positive integers, denoted as $\mathbb{N}_+$.
	For each $\phi_{i,j,k}$, output channel number is set as $c_i \cdot r_k$, where channel hyper-parameter $c_i \in \mathbb{N}_+$ and multiplication hyper-parameter $r_k \in \mathbb{N}_+$ are used to control the feature representation capabilities.
	Besides, spatial hyper-parameter $d_j \in \mathbb{N}_+$ is utilized to embed $\Phi_{j,k}$ with various spatial information, when the dilation rate of $\phi_{i,j,k}$ is set as $d_j$.
	
	As illustrated in Fig.\ref{fig:SAM}, given features $\mathbf{F}^{(1)}, \mathbf{F}^{(2)}$ deriving from $I_1,I_2$ respectively in siamese semantic segmentation branches, aSP integrates features calculated by each $\phi_{i,j,k} \in \Phi_{j,k}$.
	Then, we utilize squeeze gate $g_{j,k}$ consisting of convolutions and skip connections to reduce the computational complexity. Concretely, for $t=1,2$, we have
	\begin{equation}\label{4}
	\mathbf{F}^{(t)}_{i,j,k} =  v^{(t)}_{i,j} \cdot \phi_{i,j,k}\big(\mathbf{F}^{(t)}\big),
	\end{equation}
	\begin{equation}\label{5}
	\mathbf{F}^{(t)}_{j,k} = g_{j,k}\big(\cup^{N_c}_{i=1} \mathbf{F}^{(t)}_{i,j,k} , \,\mathbf
	{F}^{(t)}\big),
	\end{equation}
	where $\cup$ is the concatenation operator. $v^{(t)}_{i,j}$ is the element in normalized branch weights $v^{(t)} \in \mathbb{R}^{N_c \cdot N_d}$, which are obtained by global pooling and multi-layer perceptron based on a shallow layer in semantic segmentation branches. As a kind of attention weights, $v^{(t)}$ could be adjusted adaptively based on each input. The skip connection in squeeze gate $g_{j,k}$ connects $\mathbf{F}^{(t)}$ with the output to avoid gradient vanishing.
	The integrated feature map $\mathbf{F}^{(t)}_{j,k}$ embeds spatial information of various area sizes \emph{w.r.t.} each index $j$. Given each index $k$, $\{\mathbf{F}^{(1)}_{j,k}\}_{1\le j\le N_d}$ and $\{\mathbf{F}^{(2)}_{j,k}\}_{1\le j\le N_d}$ compose a pair of siamese spatial feature pyramid.
	
	Further to make receptive fields flexible and adaptive to each input during the changed region extraction, we design dense connected architectures to link $\mathbf{F}^{(1)}_{j,k}$ and $\mathbf{F}^{(2)}_{j,k}$ across siamese branches. Given index $k$, for $1 \le j_1, j_2 \le N_d$, we have
	\begin{equation}\label{6}
	\mathbf{M}_{j_1,j_2,k} = w^{(1)}_{k} \cdot \mathbf{F}^{(1)}_{j_1,k}- w^{(2)}_{k} \cdot \mathbf{F}^{(2)}_{j_2,k} \, ,
	\end{equation}
	where $w^{(1)}_k, w^{(2)}_k \in \mathbb{R}^{N_r}$ are normalized branch weights calculated from shallow layers in semantic segmentation branches. Features maps $\mathbf{F}^{(1)}_{j_1,k}$ and $\mathbf{F}^{(2)}_{j_2,k}$ are mostly generated from locally asymmetric structures and involve different sizes of image areas {\em w.r.t} each input, which makes $\{\mathbf{M}_{j_1,j_2,k}\}_{1 \le k \le N_r}$ \emph{locally asymmetric in terms of spatial information}. After the convolution operations, $\{\mathbf{M}_{j_1,j_2,k}\}_{1 \le k \le N_r}$ is passed through the decoder in change detection branch to calculate the change probability map $\mathcal{C}_{I_1, I_2}$, where $N_r$ represents the length of aSP.
	
	Besides, we fuse $\mathbf{F}^{(t)}_{j,k}$ along index $j$, saying that for $t=1,2$, 
	\begin{equation}\label{7}
	\mathbf{F}_{k}^{(t)} = h_{k}\big(w^{(t)}_{k} \cdot \cup^{N_d+1}_{j=1}\mathbf{F}^{(t)}_{j,k}, \, \mathbf{F}^{(t)}_s \big),
	\end{equation}
	where $h_{k}$ represents the squeeze gate concatenating global features, denoted as $\mathbf{F}^{(t)}_{N_d+1,k}$, with $\{\mathbf{F}^{(t)}_{j,k}\}_{j<N_d+1}$. Also, $h_{k}$ connects shallow features $\mathbf{F}^{(t)}_s$ deriving from $I_t$ to avoid gradient vanishing. In this way, $\mathbf{F}^{(t)}_{k}$ with various receptive fields is passed through the siamese decoder in siamese semantic segmentation branches to obtain semantic probability maps $\mathcal{M}_1$, $\mathcal{M}_2$. Then, semantic prediction maps $\mathcal{L_M}_1$, $\mathcal{L_M}_2$ can be obtained by selecting the semantic category with the largest probability \emph{w.r.t.} each position. 
	
	\begin{figure}[t!]
		\centering
		{\includegraphics[width=0.8\linewidth]{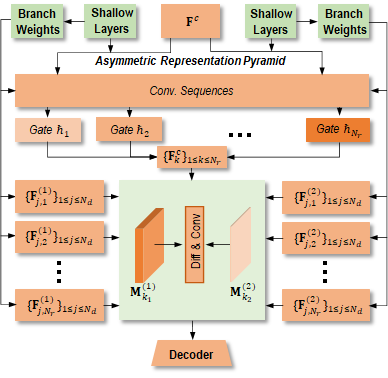}}	
		\vspace{-4mm}
		\caption{The proposed aRP. Index $k_1$, $k_2$ indicate different feature representation capabilities. Receiving spatial feature pyramids from aSP, aRP fuses asymmetric representation feature pairs with various representation capabilities.}
		\vspace{-3mm}
		\label{fig:CAM}
	\end{figure}
	
	\subsubsection{Asymmetric Representation Pyramid (aRP)}
	As illustrated in Fig.\ref{fig:CAM}, given deep features $\mathbf{F}^{c}$ in change detection branch, we leverage convolution sequences and squeeze gates to obtain $\mathbf{F}^{c}_{j,k}$ in the same way as that obtaining $\mathbf{F}^{(t)}_{j,k}$.
	Moreover, We apply normalized branch weights $w^c=(w^c_1,\cdots,w^c_{N_r}) \in \mathbb{R}^{N_r}$ to integrate $\mathbf{F}^c_{j,k}$ along the index $j$. Concretely, we have
	\begin{equation}\label{8}
	\mathbf{F}^c_{k} = \hbar_k\big( w^c_{k} \cdot \cup^{N_d+1}_{j=1}\mathbf{F}^c_{j,k}, \, \mathbf{F}^c_{s} \big) ,
	\end{equation}
	where ${\hbar_k}$ is also designed squeeze gate consisting of convolution layers and skip connections. Similarly, ${\hbar_k}$ concatenates global features, denoted as $\mathbf{F}^c_{N_d+1,k}$, with $\{\mathbf{F}^c_{j,k}\}_{j<N_d+1}$ , while ${\hbar_k}$ connects shallow features $\mathbf{F}^c_{s}$ calculated from $I_1, I_2$ with the output.
	
	As illustrated in Fig.\ref{fig:CAM}, we integrate $\mathbf{F}^c_{k}$ with $\{\mathbf{F}^{(1)}_{j,k}\}_{1 \le j \le N_d}$ and $\{\mathbf{F}^{(2)}_{j,k}\}_{1 \le j \le N_d}$ {\em{w.r.t}} each index $k$ by concatenation and convolution to obtain $\mathbf{M}_{k}^{(1)}$ and $\mathbf{M}_{k}^{(2)}$ respectively. All these $\{\mathbf{M}_{k}^{(1)}\}_{1 \le k \le N_r}, \{\mathbf{M}_{k}^{(2)}\}_{1 \le k \le N_r}$ compose a pair of siamese representation feature pyramid in aRP.
	
	Further to make feature representation capabilities flexible and adaptive to each input, we also link $\mathbf{M}_{k}^{(1)},\mathbf{M}_{k}^{(2)}$ in pairs. For $1 \le k_1,k_2 \le N_r$, we have
	\begin{equation}\label{9}
	\mathbf{M}^c_{k_1,k_2} = \mathbf{M}^{(1)}_{k_1}-\mathbf{M}^{(2)}_{k_2} .
	\end{equation}
	$\mathbf{M}^{(1)}_{k_1}, \mathbf{M}^{(2)}_{k_2}$ contain different representation capabilities \emph{w.r.t.} the index $k_1, k_2$, which make $\mathbf{M}^c_{k_1,k_2}$ implicates different representation capabilities \emph{w.r.t.} each input, namely \emph{locally asymmetric in terms of representation capabilities}.
	Finally, after convolution operations, $\{\mathbf{M}^c_{k_1,k_2}\}_{1 \le k_1, k_2 \le N_r}$ is also passed through the decoder in change detection branch to calculate the change probability map $\mathcal{C}_{I_1, I_2}$. 
	
	Given the semantic prediction maps $\mathcal{L_{M}}_1,\mathcal{L_{M}}_2$ and change probability map $\mathcal{C}_{I_1, I_2}$, we can formulate our model as follows:
	\begin{equation*}\label{9}
	\forall \mathbf p, \, f_{I_1,I_2}(\mathbf p)=\left\{\begin{array}{ll}
	(0,0),&\mathcal{C}_{I_1, I_2}(\mathbf p)<\tau,\\
	(\mathcal{L_{M}}_1(\mathbf p),\mathcal{L_{M}}_2(\mathbf p)),&
	\text{otherwise}.
	\end{array}\right.
	\end{equation*}
	
	\subsubsection{Loss Function}\label{loss}
	We utilize fully supervised learning to optimize parameters in the proposed ASN. Specifically, given the ground truth $\mathcal{G}(\mathbf p,I_1,I_2)$, which not only locates changed pixels but also indicates semantic categories, we can get ground truth for semantic probability maps ($\mathcal{L_G}_1$, $\mathcal{L_G}_2$) and change probability map $\mathcal{L_G}_c$. Then, we have
	\begin{equation}\label{10}
	\mathcal{L} = \alpha \mathcal{E}(\mathcal{M}_1, \mathcal{L_G}_1) + \beta \mathcal{E}(\mathcal{M}_2, \mathcal{L_G}_2) + \mathcal{E}(\mathcal{C}_{I_1, I_2}, \mathcal{L_G}_c),
	\end{equation}
	where $\mathcal{E}$ is the cross entropy function. $\alpha$ and $\beta$ are loss weights to adjust respective loss terms. Stochastic gradient descent (SGD) is then used to reduce the total loss and obtain the optimal parameters, through which we train all branches simultaneously.
	
	\subsubsection{Adaptive Threshold Learning}
	After the traditional training process, we design an adaptive threshold learning module to search optimal thresholds based on different outputs. Given each probability map before softmax, denoted as $\mathcal{M}^{raw}_1$, $\mathcal{M}^{raw}_2$ and $\mathcal{C}^{raw}_{I_1, I_2}$, for $t=1,2$, we have
	\begin{equation}\label{11}
	\hat{\mathcal{M}}^{raw}_t = \mathcal{M}^{raw}_t + \gamma \psi_1(\mathcal{M}^{raw}_t) ,
	\end{equation}
	\begin{equation}\label{12}
	\hat{\mathcal{C}}_{I_1, I_2} = s_{max}(\mathcal{C}^{raw}_{I_1, I_2} + \gamma \psi_2(\cup^{2}_{t=1}\hat{\mathcal{M}}^{raw}_t)) ,
	\end{equation}
	where $\psi_1, \psi_2$ are both convolution layer series with length of 2 and $s_{max}$ represents softmax function. We fix the parameters in the whole model except $\psi_1, \psi_2$ and re-train the model by utilizing the total loss in Sec.~\ref{loss} calculated based on $s_{max}(\hat{\mathcal{M}}^{raw}_t)$ and $\hat{\mathcal{C}}_{I_1, I_2}$ with categorical weights. Thus, the model can be re-formulated as 
	\begin{equation}\label{13}
	\hat{f}_{I_1,I_2}(\mathbf p)=\left\{\begin{array}{ll}
	(0,0),&\hat{\mathcal{C}}_{I_1, I_2}(\mathbf p)<\tau,\\
	(\mathcal{L_{\hat{M}}}_1(\mathbf p),\mathcal{L_{\hat{M}}}_2(\mathbf p)),&
	\text{otherwise} .
	\end{array}\right.
	\end{equation}
	$\mathcal{L_{\hat{M}}}_1(\mathbf p),\mathcal{L_{\hat{M}}}_2(\mathbf p)$ are the updated semantic prediction maps obtained based on $s_{max}(\hat{\mathcal{M}}^{raw}_1), s_{max}(\hat{\mathcal{M}}^{raw}_2)$ respectively.
	During the parameter optimization, ATL explores the logical relationships between semantic categories within considered neighbourhood, which can adaptively update the semantic prediction maps and revise the threshold deflections caused by label imbalance.
	
	\section{The SECOND Dataset}
	Although several change detection datasets have been proposed~\cite{HRSCD, Hermiston_City_area, quasi_urban_areas, OSCD, ACD, SemiCDNet}, only few of them contain land-cover categorical information, which is needed for SCD. A natural way to create an SCD dataset is comparing multi-temporal land-cover maps in the same geographic locations \cite{HRSCD}, which would neglect change types such as the demolition and reconstruction of \emph{buildings}. Moreover, due to the required dense labors, a large-scale SCD dataset is hard to acquire, while the annotations of land-cover classes also require professional knowledge.
	
	In order to set up a new benchmark for SCD problems with adequate quantities, sufficient categories and proper annotation methods, in this paper we present SECOND, a well-annotated semantic change detection dataset. Different from the datasets used in ~\cite{Mou_TGRS, Innovative_Neural_Net_Method_tgrs}, to ensure data diversity, we firstly collect 4662 pairs of aerial images from several platforms and sensors. These pairs of images are distributed over the cities such as Hangzhou, Chengdu, and Shanghai. Each image has size 512 $\times$ 512 and is annotated at the pixel level. The annotation of SECOND is carried out by an expert group of earth vision applications, which guarantees high label accuracy.
	\begin{figure*}
		\centering
		{\includegraphics[width=0.95\linewidth]{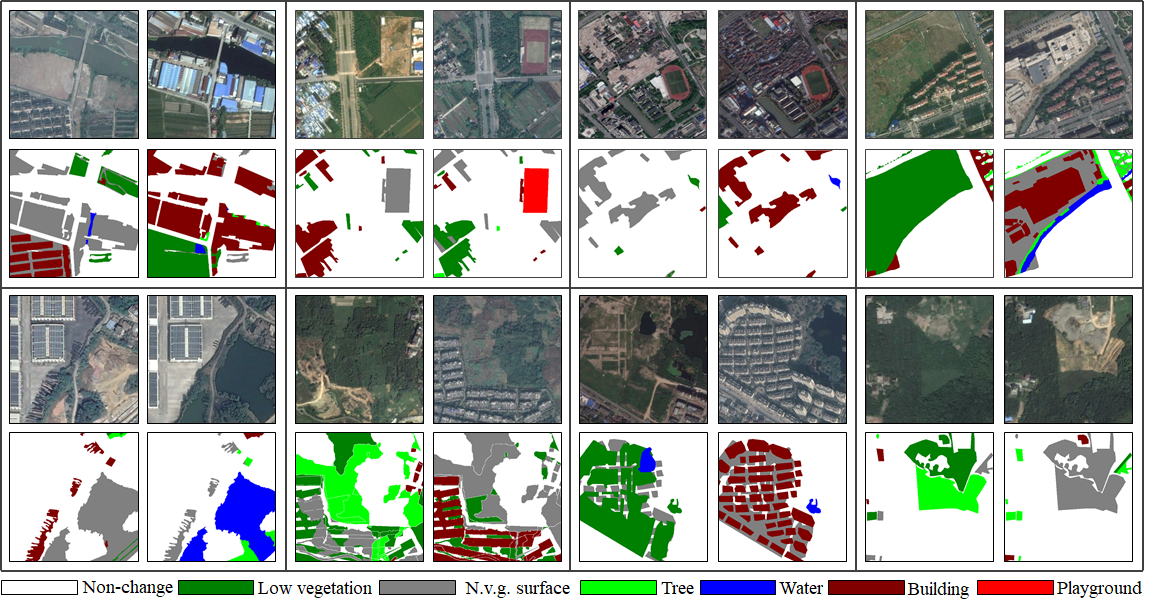}}	
		\vspace{-3mm}
		\caption{Several samples of our proposed SECOND dataset. Color white indicates \emph{non-change} regions, while other colors indicate different land-cover classes. Ground truth for SCD can be obtained by comparing the annotated land-cover classes.}
		\vspace{-3mm}
		\label{fig:Dataset_samples}
	\end{figure*}
	\begin{figure*}
		\centering
		{\includegraphics[width=1.0\linewidth]{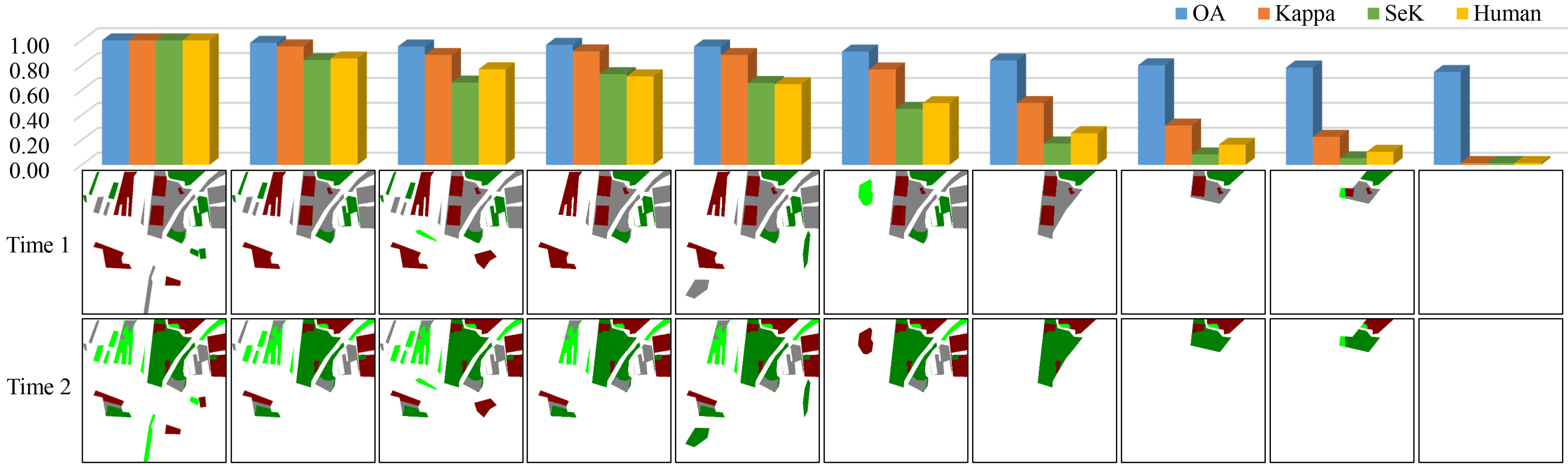}}
		\vspace{-7mm}
		\caption{Comparison between SeK and other metrics. SeK alleviates the influence of label imbalance caused by \emph{non-change} pixels and shows most similarity with human scoring, while outputs with dominant predictions of \emph{non-change} get unreasonable high scores in OA and $\kappa$.}
		\vspace{-2mm}
		\label{fig:Label}
	\end{figure*}
	Moreover, in contrast with \cite{HRSCD}, we utilize land-cover map pairs and \emph{non-change} masks to represent the change types. The \emph{non-change} masks equal the ground truth for BCD and are used to blacken the unchanged pixels in siamese semantic segmentation branches during the training process, which can avoid semantic categorical ambiguity. Through this annotation method, we can distinguish \emph{non-change} pixels from changed pixels between the same land-cover class.
	
	For the change type in the SECOND dataset, we focus on 6 main land-cover classes, {\em i.e.}, {\em non-vegetated ground surface, tree, low vegetation, water, buildings} and \emph{playgrounds}, that are frequently involved in natural and man-made geographical changes \cite{1983Estimation,2020Land,DiRs,DOTA}. 
	It is worth noticing that, in the new dataset, non-vegetated ground surface  ({\em n.v.g. surface} for short) mainly corresponds to \emph{impervious surface} and \emph{bare land}.
	In summary, these 6 selected land-cover classes result in 30 common change types (including \emph{non-change}).
	Through the random selection of image pairs, the SECOND reflects real distributions of land-cover classes when changes occur.
	Several samples of SECOND are displayed in Fig.\ref{fig:Dataset_samples}, where we can see the data diversity and label accuracy. The SECOND dataset is available at \url{http://www.captain-whu.com/project/SCD}.
	
	\section{Evaluation Metrics}
	\subsection{Overall accuracy and Kappa coefficient}
	Existing works often utilize Overall Accuracy (OA) and Kappa coefficient ($\kappa$) that are commonly used for measuring classification performance to evaluate change detection algorithms~\cite{HRSCD, Mou_TGRS}. Given a confusion matrix $Q=\{q_{ij}\}$, where $q_{ij}$ indicates the number of pixels that are identified as the $i$-{th} change type and actually belong to the $j$-th change type (\emph{non-change} is set as the first change type), then OA is defined as
	\begin{align}\label{14}
	\textrm{OA} \triangleq \rho &=\sum^C_{i=1}{q_{ii}}/\sum^C_{i=1}\sum\limits^C_{j=1}{q_{ij}}, 
	\end{align}
	where $C$ is the total number of change types. Due to the equivalence of each pixel in the calculation of OA, dominant \emph{non-change} pixels would cause unreasonable scores. While, as a statistic calculated from confusion matrix (a kind of contingency table), $\kappa$ measures the consistency between outputs and labels, which is less affected by the label imbalance \cite{Cohen1960A}. More precisely, we have
	\begin{align}
	\kappa =(\rho-\eta)/(1-\eta),
	\end{align}
	where $\eta =\sum^C_{j=1}{(q_{j+} \cdot q_{+j})}/(\sum^C_{i=1}\sum^C_{j=1}{q_{ij}})^{2}$ with $q_{j+}$ and $q_{+j}$ being as the row sum and column sum of the confusion matrix $Q$. 
	
	However, the dominant \emph{non-change} pixels still mislead the scores obtained by $\kappa$.
	Given a change detection data sample, {\em{i.e.}} a pair of images and a sequence of change detection results, we collect visual scores between 0 and 1 \emph{w.r.t.} each result from 11 remote sensing image interpretation experts. Meanwhile, we calculate evaluation scores of each result based on OA and $\kappa$.
	As illustrated in Fig.~\ref{fig:Label}, the collapse model with constant \emph{non-change} predictions would get unreasonable high scores in OA. Besides, although $\kappa$ tends to zero when the model gradually collapses, the score of second to last result still gets 0.23 in $\kappa$, which is too high compared with human scoring.
	
	
	\begin{figure*}
		\centering
		{\includegraphics[width=0.94\linewidth]{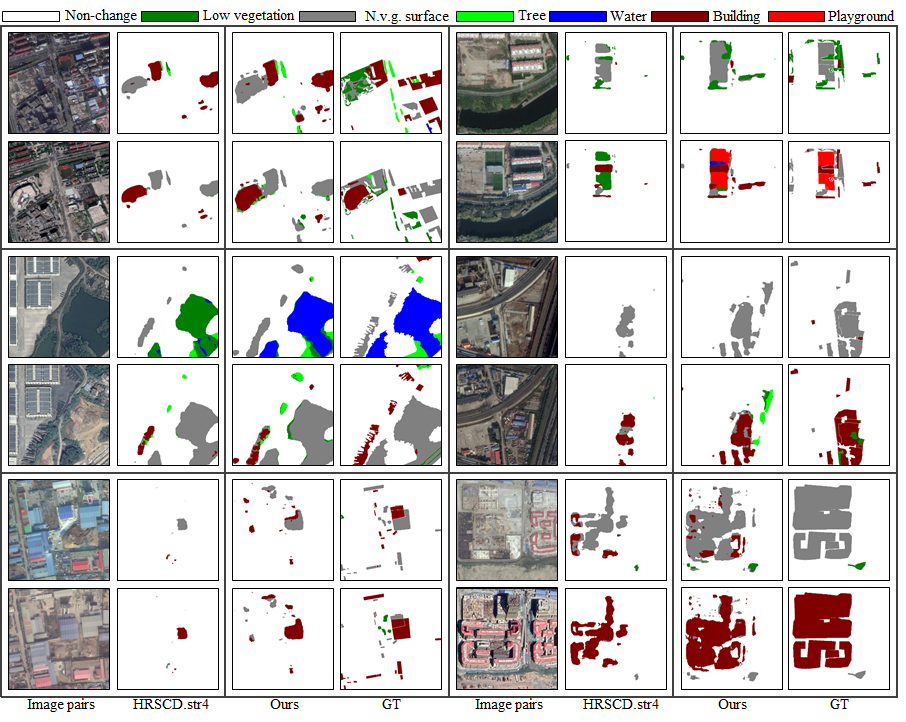}}	
		\vspace{-3mm}
		\caption{Visual results of comparison with state-of-the-art method when the encoder is built on residual blocks. We mask the semantic prediction maps with change maps to represent the prediction of change type in each position, where our proposed ASN can better identify land-cover classes and alleviate false identifications of \emph{non-change} pixels.}
		\label{fig:Residual}
		\vspace{-2mm}
	\end{figure*}
	
	\subsection{Separated Kappa (SeK) coefficient}
	In order to alleviate the influence of label imbalance, we utilize mIOU to evaluate BCD results and propose a SeK coefficient to evaluate SCD results. Specifically, given a confusion matrix Q, we have
	\begin{align}
	\label{16}
	\textrm{IOU}_1 &= q_{11}/(\sum\limits^C_{i=1}{q_{i1}}+\sum\limits^C_{j=1}{q_{1j}}-q_{11}), \\
	\label{17}
	\textrm{IOU}_2 &= \sum\limits^C_{i=2}\sum\limits^C_{j=2}{q_{ij}}/(\sum\limits^C_{i=1}\sum\limits^C_{j=1}{q_{ij}}-q_{11}) ,
	\end{align}	
	where $\textrm{IOU}_1$ measures the identification of \emph{non-change} pixels and $\textrm{IOU}_2$ evaluates the extraction of changed regions.
	Then, we have
	\begin{align}\label{20}
	\textrm{mIOU}=\frac{1}{2}(\textrm{IOU}_1+\textrm{IOU}_2).
	\end{align} 
	Involving in $\textrm{IOU}_2$, mIOU considers more about changed regions. 
	
	On the other hand, the true positive of \emph{non-change} pixels $q_{11}$ always dominates the calculation of $\kappa$. Thus, we separate $q_{11}$ in the calculation of SeK. We also utilize $\textrm{IOU}_2$ to further emphasize changed pixels.
	Specifically, we define
	\begin{align}\label{20}
	\textrm{SeK} =e^{(\textrm{IOU}_2-1)} \cdot (\hat{\rho}-\hat{\eta})/(1-\hat{\eta}),
	\end{align} 	
	with 
	\begin{align}
	\hat{\rho} &=\sum^C_{i=2}{q_{ii}}/(\sum\limits^C_{i=1}\sum^C_{j=1}{q_{ij}}-q_{11}), \nonumber\\
	\hat{\eta} &=\sum^C_{j=1}{(\hat{q}_{j+} \cdot \hat{q}_{+j})}/(\sum^C_{i=1}\sum^C_{j=1}{q_{ij}}-q_{11})^{2}, \nonumber
	\end{align} 
	where $\hat{q}_{j+}$ and $\hat{q}_{+j}$ represent the row sum and column sum of confusion matrix without $q_{11}$. The exponential form enlarges the discernibility compared with simple multiplication when evaluating models with better performance. 
	
	As illustrated in Fig.\ref{fig:Label}, compared with $\kappa$ and OA, models with apparently poor performances on small change types would get low scores in SeK no matter how good the performances on BCD are. Moreover, the Mean Square Error (MSE) between SeK and human scores is 0.003. While, MSE \emph{w.r.t.} OA and $\kappa$ are 0.212 and 0.028 respectively, which further validates the rationality of SeK. 
	
	
	
	\section{Experiments and Analysis}
	\label{sec:Experiments}	
	In this section, we evaluate the proposed ASN on the SECOND dataset. We first clarify the experiment settings in Sec.~\ref{Exp_Settings}. Then, in Sec.~\ref{Basic Architecture Analysis} and Sec.~\ref{STOA}, we discuss several existing structures and compare ASN with several methods, including natural extensions of classical BCD algorithms and the state-of-the-art SCD algorithms~\cite{HRSCD, CD_FULLY_ICIP} using backbones with different basic blocks~\cite{Resnet,Xception,SENet}. Further in Sec.~\ref{FV}, we demonstrate the effectiveness of our proposed modules by feature visualization. Finally, in Sec.~\ref{Ablation}, we remove the aSP, aRP and asymmetric feature pairs to make comparison with original ASN and verify the merits of each term in the proposed model.
	
	\subsection{Experiment Settings}\label{Exp_Settings}
	The change detection algorithms involved in the comparison experiments are as follows:
	\begin{itemize}
		\item[-] FC-EF\cite{CD_FULLY_ICIP}: a BCD algorithm using single encoder-decoder structure.
		\item[-] FC-conc\cite{CD_FULLY_ICIP}: a BCD algorithm using siamese encoders followed with single decoder branch and concatenation skip connections from the encoder to the decoder.
		\item[-] FC-diff\cite{CD_FULLY_ICIP}: a BCD algorithm using siamese encoders followed with single decoder branch and difference skip connections from the encoder to the decoder.
		\item[-] HRSCD.str1\cite{HRSCD}: an algorithm corresponding to the intuitive solution to SCD problem discussed in Sec.~\ref{sec:Intuitive}.
		\item[-] HRSCD.str2\cite{HRSCD}: a SCD algorithm using single encoder-decoder structure.
		\item[-] HRSCD.str3\cite{HRSCD}: a SCD algorithm using siamese semantic segmentation branches with change detection branch.
		\item[-] HRSCD.str4\cite{HRSCD}: a SCD algorithm using siamese semantic segmentation branches with change detection branch and difference skip connections from siamese encoders to the decoder of change detection branch.
	\end{itemize}
	
	
	\begin{table}[htb!]
		\scriptsize
		\centering
		\renewcommand\tabcolsep{3.9pt} 
		\small
		\caption{Comparison with state-of-the-art methods when the encoder is built on residual blocks. }
		\vspace{-2mm}
		\label{tab:Overall-Comparison-Residual}
		\setlength{\arraycolsep}{0.1pt}
		\begin{threeparttable}
			\begin{tabular}{c|cc|cc|cc}
				
				\toprule
				
				\multirow{2}{*}{Methods} & \multicolumn{2}{c}{MS \underline{\space\space\space} Flip \underline{\space\space\space}} & \multicolumn{2}{c}{MS \underline{\checkmark} Flip \underline{\space\space\space}} & \multicolumn{2}{c}{MS \underline{\checkmark} Flip \underline{\checkmark} } \\
				
				\cmidrule(r){2-3} \cmidrule(r){4-5} \cmidrule(r){6-7}
				
				&mIOU&SeK&mIOU&SeK&mIOU&SeK\\
				
				\midrule
				
				FC-EF~\cite{CD_FULLY_ICIP}&59.3&5.7&59.3&5.7&59.0&5.9\\
				FC-conc~\cite{CD_FULLY_ICIP}&63.3&9.1&62.8&9.2&62.9&9.4\\
				FC-diff~\cite{CD_FULLY_ICIP}&61.9&8.8&60.9&8.6&61.0&8.7\\
				HRSCD.str1~\cite{HRSCD}&29.3&4.6&29.8&4.9&29.8&4.9\\
				HRSCD.str2~\cite{HRSCD}&59.7&6.3&59.4&6.5&59.4&6.6\\			
				HRSCD.str3~\cite{HRSCD}&62.3&8.9&62.0&9.1&62.1&9.2\\
				HRSCD.str4~\cite{HRSCD}&67.5&13.7&67.8&14.4&67.9&14.5\\
				\hline
				ASN (Ours)&69.0&15.2&69.5&16.1&69.7&16.2\\	
				ASN-ATL (Ours)&\textbf{69.1}&\textbf{15.5}&\textbf{69.8}&\textbf{16.5}&\textbf{70.0}&\textbf{16.8}\\	
				\bottomrule
			\end{tabular}
		\end{threeparttable}
	\end{table}
	\begin{table*}[htb!]
		\scriptsize
		\centering
		\renewcommand\tabcolsep{4.7pt} 
		\caption{Detail categorical results with three kinds of testing strategies when the encoder is built on residual blocks. Top sheets are without testing strategy. Middle sheets are with MS testing strategy. Bottom sheets are with MS and Flip testing strategy. The categorical SeK is listed in the matrices, while the categorical intersection over union ($\textrm{IOU}_1,\textrm{IOU}_2$) of binary change detection is listed below each matrices.}
		\vspace{-2mm}
		\small
		\label{Tab-Detail-Residual}
		\setlength{\arraycolsep}{0.1pt}
		\begin{tabular}{c|cccccc|cccccc|cccccc}
			
			\toprule
			
			\multirow{6}{*}{Residual} & \multicolumn{6}{c}{HRSCD.str1(intuitive solution)} & \multicolumn{6}{c}{HRSCD.str4} & \multicolumn{6}{c}{ASN-ATL (Ours)} \\
			
			\cmidrule(r){2-7} \cmidrule(r){8-13} \cmidrule(r){14-19}
			
			&\rotatebox{90}{Low vegetation}&\rotatebox{90}{N.v.g. surface}&\rotatebox{90}{Building}&\rotatebox{90}{Tree}&\rotatebox{90}{Water}&\rotatebox{90}{Playground}&\rotatebox{90}{Low vegetation}&\rotatebox{90}{N.v.g. surface}&\rotatebox{90}{Building}&\rotatebox{90}{Tree}&\rotatebox{90}{Water}&\rotatebox{90}{Playground}&\rotatebox{90}{Low vegetation}&\rotatebox{90}{N.v.g. surface}&\rotatebox{90}{Building}&\rotatebox{90}{Tree}&\rotatebox{90}{Water}&\rotatebox{90}{Playground}\\
			
			\midrule
			
			Low vegetation&--&30.0&22.7&19.4&21.7&3.3&--&41.7&30.8&22.2&19.9&8.4&--&\textbf{42.0}&\textbf{32.2}&\textbf{28.3}&\textbf{20.4}&\textbf{34.2}\\
			
			N.v.g. surface&31.3&--&31.3&20.7&15.0&4.7&\textbf{44.1}&--&42.1&28.5&24.6&6.0&43.7&--&\textbf{44.7}&\textbf{30.4}&\textbf{29.5}&\textbf{41.9}\\
			
			Building&21.9&31.3&21.4&12.6&19.2&0.0&\textbf{35.1}&47.2&26.1&18.1&\textbf{33.8}&0.0&33.1&\textbf{48.3}&\textbf{27.1}&\textbf{24.9}&31.1&\textbf{17.8}\\
			
			Tree&16.9&19.1&16.1&--&1.1&0.0&12.0&26.0&\textbf{24.5}&--&\textbf{12.6}&0.0&\textbf{20.2}&\textbf{27.7}&23.7&--&5.9&0.0\\
			Water&19.5&16.1&19.0&10.0&--&--&13.7&26.0&\textbf{29.2}&13.0&--&--&\textbf{25.9}&\textbf{29.8}&27.8&\textbf{21.7}&--&--\\
			Playground&0.1&3.4&0.0&0.0&--&--&1.1&1.8&0.4&0.0&--&--&\textbf{17.1}&\textbf{29.7}&\textbf{16.5}&\textbf{1.3}&--&--\\	
			\hline
			Non-change& \multicolumn{6}{c}{$\textrm{IOU}_1:34.9$ \space\space\space\space\space\space\space $\textrm{IOU}_2:23.8$} & \multicolumn{6}{c}{$\textrm{IOU}_1:\textbf{86.6}$ \space\space\space\space\space\space\space $\textrm{IOU}_2:48.4$} & \multicolumn{6}{c}{$\textrm{IOU}_1:$85.9 \space\space\space\space\space\space\space $\textrm{IOU}_2:$\textbf{52.4}}\\
			\hline
			Low vegetation&--&30.6&23.1&20.0&\textbf{22.0}&1.2&--&42.4&31.6&21.8&14.4&4.7&--&\textbf{43.4}&\textbf{32.5}&\textbf{28.5}&21.2&\textbf{37.6}\\
			
			N.v.g. surface&31.5&--&31.9&21.4&14.3&1.7&44.7&--&42.8&29.2&19.8&4.8&\textbf{45.4}&--&\textbf{45.4}&\textbf{31.5}&\textbf{31.5}&\textbf{43.3}\\
			
			Building&22.4&32.3&22.5&13.9&20.5&0.0&\textbf{36.1}&48.3&27.9&21.4&31.2&0.0&34.4&\textbf{49.4}&\textbf{28.7}&\textbf{25.9}&\textbf{36.0}&\textbf{20.1}\\
			
			Tree&17.8&19.8&16.9&--&0.9&0.0&12.0&26.8&\textbf{25.6}&--&7.8&0.0&\textbf{21.0}&\textbf{28.7}&24.8&--&\textbf{10.3}&0.0\\
			Water&16.4&15.9&17.8&11.0&--&--&7.5&22.0&26.3&13.0&--&--&\textbf{26.8}&\textbf{31.4}&\textbf{28.8}&\textbf{21.3}&--&--\\
			Playground&0.0&1.9&0.0&0.0&--&--&0.1&1.7&0.3&0.0&--&--&\textbf{15.5}&\textbf{28.0}&\textbf{20.8}&\textbf{0.1}&--&--\\	
			\hline
			Non-change& \multicolumn{6}{c}{$\textrm{IOU}_1:35.6$ \space\space\space\space\space\space\space $\textrm{IOU}_2:24.0$} & \multicolumn{6}{c}{$\textrm{IOU}_1:\textbf{86.9}$ \space\space\space\space\space\space\space $\textrm{IOU}_2:48.8$} & \multicolumn{6}{c}{$\textrm{IOU}_1:86.4$ \space\space\space\space\space\space\space $\textrm{IOU}_2:\textbf{53.3}$}\\
			\hline
			Low vegetation&--&30.8&23.2&20.1&\textbf{21.9}&1.2&--&42.6&31.8&22.2&11.3&3.5&--&\textbf{43.6}&\textbf{32.7}&\textbf{28.3}&21.5&\textbf{37.7}\\
			
			N.v.g. surface&31.6&--&31.9&21.4&14.1&2.3&45.0&--&42.9&29.4&18.9&4.4&\textbf{45.7}&--&\textbf{45.7}&\textbf{31.8}&\textbf{32.5}&\textbf{43.3}\\
			
			Building&22.5&32.4&22.4&14.2&20.5&0.0&\textbf{36.3}&48.4&28.2&22.0&29.3&0.0&34.6&\textbf{49.7}&\textbf{29.1}&\textbf{26.4}&\textbf{35.5}&\textbf{20.1}\\
			
			Tree&18.0&20.0&16.8&--&0.8&0.0&12.8&26.7&\textbf{26.2}&--&6.4&0.0&\textbf{21.3}&\textbf{28.6}&25.2&--&\textbf{13.4}&0.0\\
			Water&16.1&15.4&18.1&9.6&--&--&8.7&22.2&25.7&15.4&--&--&\textbf{27.4}&\textbf{32.1}&\textbf{29.2}&\textbf{22.0}&--&--\\
			Playground&0.0&2.4&0.0&0.0&--&--&0.0&1.6&0.3&0.0&--&--&\textbf{15.9}&\textbf{27.9}&\textbf{24.9}&\textbf{0.3}&--&--\\	
			\hline
			Non-change& \multicolumn{6}{c}{$\textrm{IOU}_1:35.6$ \space\space\space\space\space\space\space $\textrm{IOU}_2:24.0$} & \multicolumn{6}{c}{$\textrm{IOU}_1:\textbf{87.0}$ \space\space\space\space\space\space\space $\textrm{IOU}_2:48.8$} & \multicolumn{6}{c}{$\textrm{IOU}_1:86.5$ \space\space\space\space\space\space\space $\textrm{IOU}_2:\textbf{53.5}$}\\ 	
			\bottomrule
		\end{tabular}
	\end{table*}
	
	\begin{table}[htb!]
		\scriptsize
		\centering
		\renewcommand\tabcolsep{3.9pt} 
		\small
		\caption{Comparison with state-of-the-art methods when the encoder is built on Xception blocks.}
		\vspace{-3mm}
		\label{tab:Overall-Comparison-Xception}
		\setlength{\arraycolsep}{0.1pt}
		\begin{threeparttable}
			\begin{tabular}{c|cc|cc|cc}
				\toprule
				
				\multirow{2}{*}{Methods} & \multicolumn{2}{c}{MS \underline{\space\space\space} Flip \underline{\space\space\space}} & \multicolumn{2}{c}{MS \underline{\checkmark} Flip \underline{\space\space\space}} & \multicolumn{2}{c}{MS \underline{\checkmark} Flip \underline{\checkmark} } \\
				
				\cmidrule(r){2-3} \cmidrule(r){4-5} \cmidrule(r){6-7}
				
				&mIOU&SeK&mIOU&SeK&mIOU&SeK\\
				
				\midrule
				
				FC-EF~\cite{CD_FULLY_ICIP}&56.2&3.6&55.6&3.9&55.7&4.0\\
				FC-conc~\cite{CD_FULLY_ICIP}&61.1&7.3&60.4&7.4&60.5&7.5\\
				FC-diff~\cite{CD_FULLY_ICIP}&57.3&5.0&56.6&5.0&56.5&5.0\\
				HRSCD.str1~\cite{HRSCD}&29.3&4.7&30.1&4.9&30.2&5.0\\
				HRSCD.str2~\cite{HRSCD}&59.7&5.7&59.0&5.8&59.0&5.9\\		
				HRSCD.str3~\cite{HRSCD}&62.1&8.4&61.9&8.7&61.9&8.8\\
				HRSCD.str4~\cite{HRSCD}&67.2&13.0&67.8&13.8&68.0&14.1\\
				\hline
				ASN (Ours)&68.4&14.3&68.8&15.2&69.0&15.5\\
				ASN-ATL (Ours)&\textbf{69.0}&\textbf{14.9}&\textbf{69.6}&\textbf{15.9}&\textbf{69.9}&\textbf{16.3}\\	
				\bottomrule
			\end{tabular}
		\end{threeparttable}
	\end{table}
	
	\begin{table}[htb!]
		\scriptsize
		\centering
		\renewcommand\tabcolsep{3.9pt} 
		\small
		\caption{Comparison with state-of-the-art methods when the encoder is built on Squeeze-and-Excitation blocks.}
		\vspace{-3mm}
		\label{tab:Overall-Comparison-SENet}
		\setlength{\arraycolsep}{0.1pt}
		\begin{threeparttable}
			\begin{tabular}{c|cc|cc|cc}
				
				\toprule
				
				\multirow{2}{*}{Methods} & \multicolumn{2}{c}{MS \underline{\space\space\space} Flip \underline{\space\space\space}} & \multicolumn{2}{c}{MS \underline{\checkmark} Flip \underline{\space\space\space}} & \multicolumn{2}{c}{MS \underline{\checkmark} Flip \underline{\checkmark} } \\
				
				\cmidrule(r){2-3} \cmidrule(r){4-5} \cmidrule(r){6-7}
				
				&mIOU&SeK&mIOU&SeK&mIOU&SeK\\
				
				\midrule
				
				FC-EF~\cite{CD_FULLY_ICIP}&60.5&6.9&60.1&6.9&60.1&6.9\\
				FC-conc~\cite{CD_FULLY_ICIP}&64.2&10.5&63.7&10.5&63.6&10.5\\
				FC-diff~\cite{CD_FULLY_ICIP}&63.1&9.7&62.3&9.6&62.3&9.7\\
				HRSCD.str1~\cite{HRSCD}&30.4&5.0&31.0&5.3&31.1&5.3\\
				HRSCD.str2~\cite{HRSCD}&62.3&8.3&61.9&8.3&61.9&8.3\\		
				HRSCD.str3~\cite{HRSCD}&62.0&8.6&62.0&9.0&62.0&9.1\\
				HRSCD.str4~\cite{HRSCD}&67.9&14.6&68.1&15.3&68.2&15.4\\
				\hline
				ASN (Ours)&69.4&15.9&69.6&16.5&69.7&16.6\\	
				ASN-ATL (Ours)&\textbf{69.5}&\textbf{16.3}&\textbf{70.1}&\textbf{17.2}&\textbf{70.2}&\textbf{17.3}\\	
				\bottomrule
			\end{tabular}
		\end{threeparttable}
	\end{table}
	
	ASN keeps the same basic architecture as HRSCD.str4, which contains 10 basic blocks in encoders and 13 basic blocks in decoders. We replace the difference skip connections with simple summations in ASN to reduce model sizes. The proposed aSP and aRP are embedded between the $6^{th}$ and $7^{th}$ basic blocks in the decoder branches.
	ASN-ATL applys adaptive threshold learning module on ASN. All the experiments are implemented on 4 Pascal V100 with memory of 16G based on Pytorch. All evaluated models are trained and tested under the same conditions without pre-trained models and other post-processings. 
	
	\begin{figure*}[!t]
		\centering
		{\includegraphics[width=0.96\linewidth]{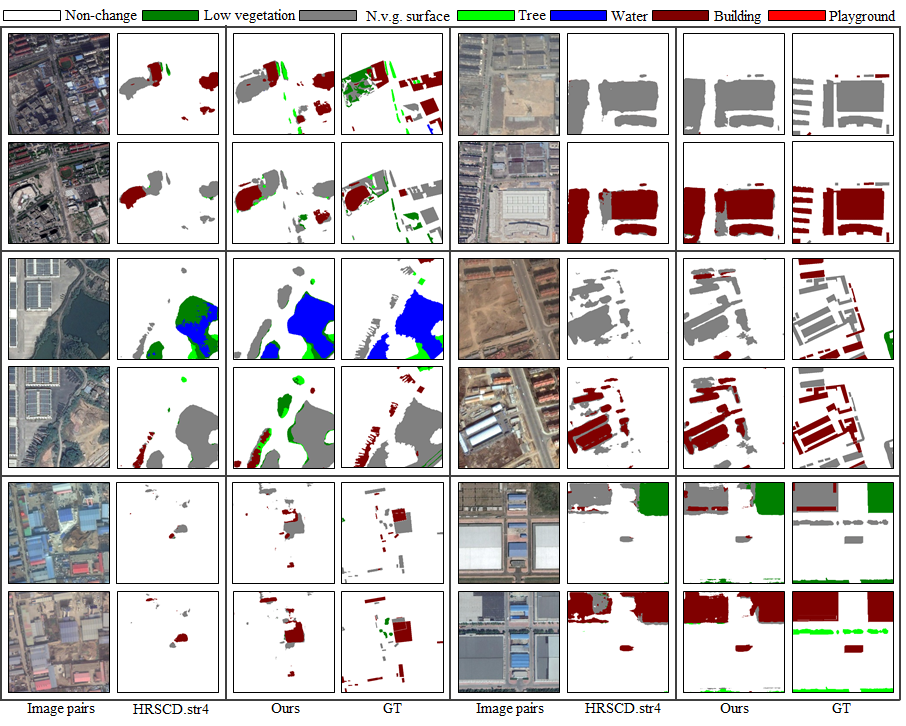}}	
		\vspace{-3mm}
		\caption{Comparisons with state-of-the-art method when the encoder is built on Xception blocks. Left part of the figure shows the stable outperformance of ASN with different backbone, while the right part exhibits the generalization ability on different data samples.}
		\label{fig:Xception}
	\end{figure*}
	\begin{figure*}[!t]
		\centering
		{\includegraphics[width=0.99\linewidth]{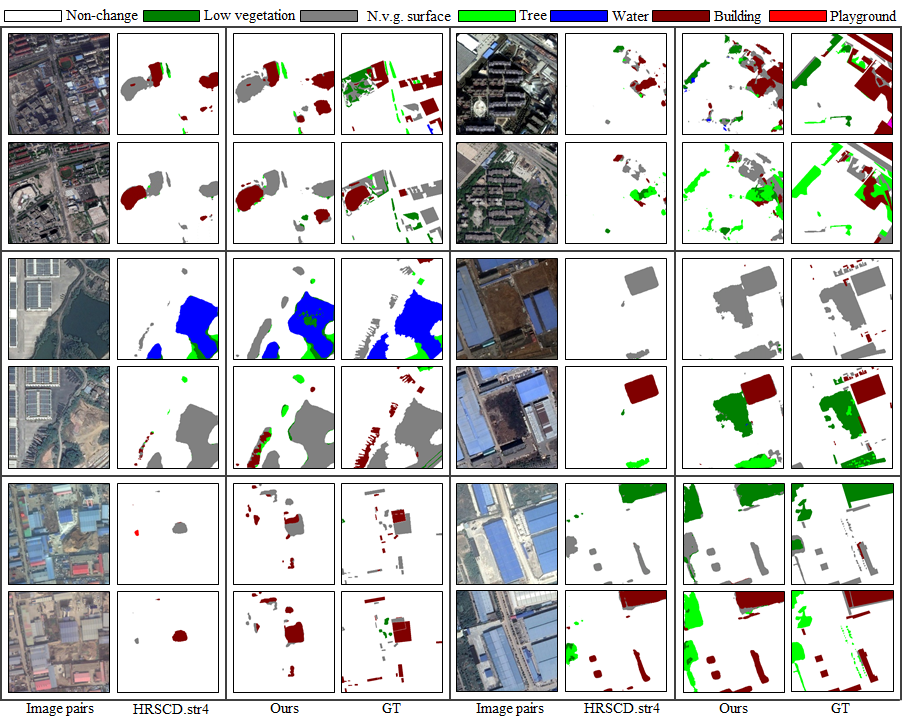}}	
		\vspace{-3mm}
		\caption{Comparisons with state-of-the-art method when the encoder is built on Squeeze-and-Excitation blocks. ASN also shows stable outperformance and generalization ability.}
		\label{fig:SENet}
	\end{figure*}
	\begin{figure*}[htb!]
		\centering
		{\includegraphics[width=0.75\linewidth]{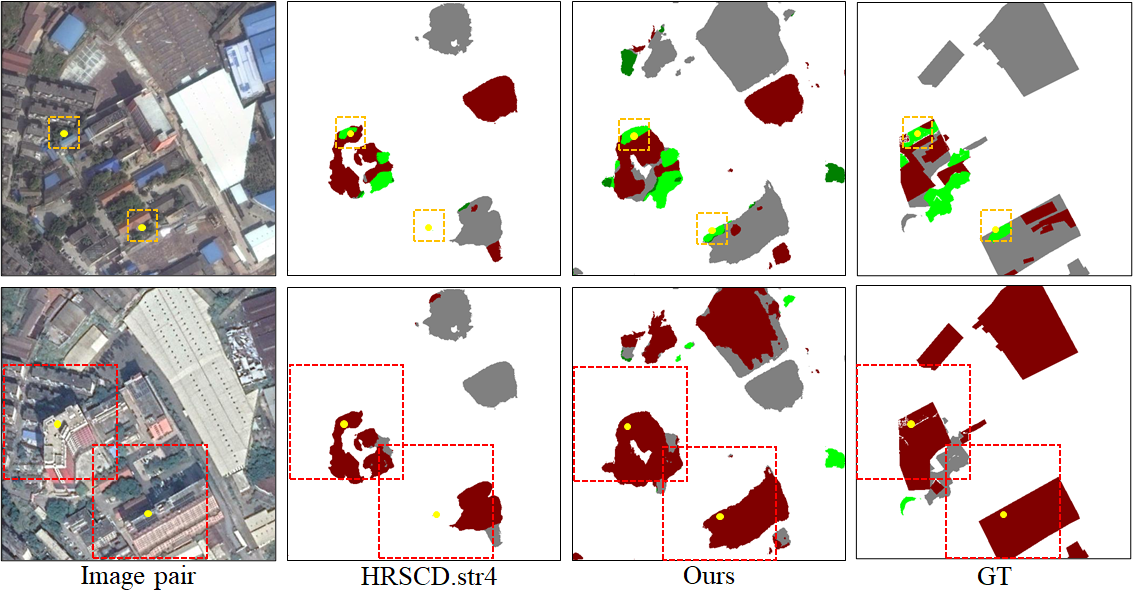}}
		\vspace{-3mm}
		\caption{Result visualizations of the case discussed in Sec.~\ref{sec:introduction} when the encoder is built on residual blocks. ASN can extract changed regions and identify change types more precisely, while addressing these asymmetric changes.}
		\label{fig:echo}
	\end{figure*}	
	
	In the training process, SGD is utilized to search optimal parameters for 50 epochs, while extra parameters in ASN-ATL would be trained for 20 epochs with other parameters fixed. Random flip and random scale between 0.5 and 2 are utilized as the data augmentation.
	The initial learning rate is set as 0.005 and 'poly' policy is employed with the power of 0.9. Batch size is set as 4. Also, the momentum is set as 0.9 and weight decay is set as 0.0001. 
	In the testing process, we apply flip strategy and multi-scale (MS) testing with 6 scales which are 0.5, 0.75, 1.0, 1.25, 1.5 and 1.75. 
	
	As for the hyper-parameters in ASN, considering in the image size in SECOND, we set spatial hyper-parameters $\{d_1,\cdots,d_{N_d}\}$ as $\{0, 6, 12\}$ to make feature receptive fields in aSP vary from local to the whole image. Besides, multiplication hyper-parameters $\{r_1,\cdots,r_{N_r}\}$ and channel hyper-parameters $\{c_1,\cdots,c_{N_c}\}$ are set as $\{16, 32, 64\}$ and $\{1, 2, 3, 4, 5\}$ to make features in aSP and aRP contain balanced representation capabilities with features in decoders. Then, the kernel size of convolution layers in aSP and aRP are all set as 3. We further set scalar threshold $\tau$ as 0.5. The hyper-parameter $\gamma$ in ATL module is also set as 0.5, while we slightly adjust it to 0.4 for Xception backbone. For all the experiments, we split SECOND dataset into two subsets: 2968 sample pairs for training and 1694 sample pairs for testing. Finally, we set $\alpha$ and $\beta$ in total loss as 1 in the training process. 
	
	
	\subsection{Discussions on Basic Structures}\label{Basic Architecture Analysis}
	As illustrated in Tab.~\ref{tab:Overall-Comparison-Residual}, HRSCD.str1 gets 29.8 in mIOU with full testing strategies, while FC-conc and FC-diff achieve 62.9 and 61.0 respectively. This result demonstrates that separate semantic segmentation application and false categorical independence assumption would limit the model performance, which could be addressed by the joint training of siamese network. Importantly, although FC-EF outperforms HRSCD.str1 dramatically in mIOU, they share similar results in SeK, which implicates that SeK would not simply increase merely with good BCD performance. The outperformances of FC-conc and FC-diff compared with FC-EF indicates that the siamese networks provide extra improvements compared with single encoder-decoder structure.
	Moreover, the outperformances of HRSCD.str3 and HRSCD.str4 show the superiority of separate changed region extraction and change type identification.

	\subsection{Comparison with State-of-the-art Algorithms}\label{STOA}
	As shown in Tab.\ref{tab:Overall-Comparison-Residual}, ASN-ATL achieves the best results, 69.1 in mIOU and 15.5 in SeK, without testing strategies. Then, with multi-scale strategy, ASN-ATL achieves 69.8 in mIOU and 16.5 in SeK, while the flip strategy provides extra 0.2 improvements in mIOU and 0.3 improvements in SeK. Our proposed model stably outperforms existing state-of-the-art algorithms, which achieves 2.1 improvements in mIOU and 2.3 improvements in SeK compared to HRSCD.str4 with full testing strategies. Also, as shown in Tab.\ref{tab:Overall-Comparison-Residual}, ATL provides 0.3 improvement in mIOU and 0.6 improvement in SeK with full testing strategies. We can also see in Fig.~\ref{fig:Residual}, the proposed model can recover more details which leads to the better performance in SeK. 
	\begin{figure*}[!t]
		\centering
		{\includegraphics[width=0.95\linewidth]{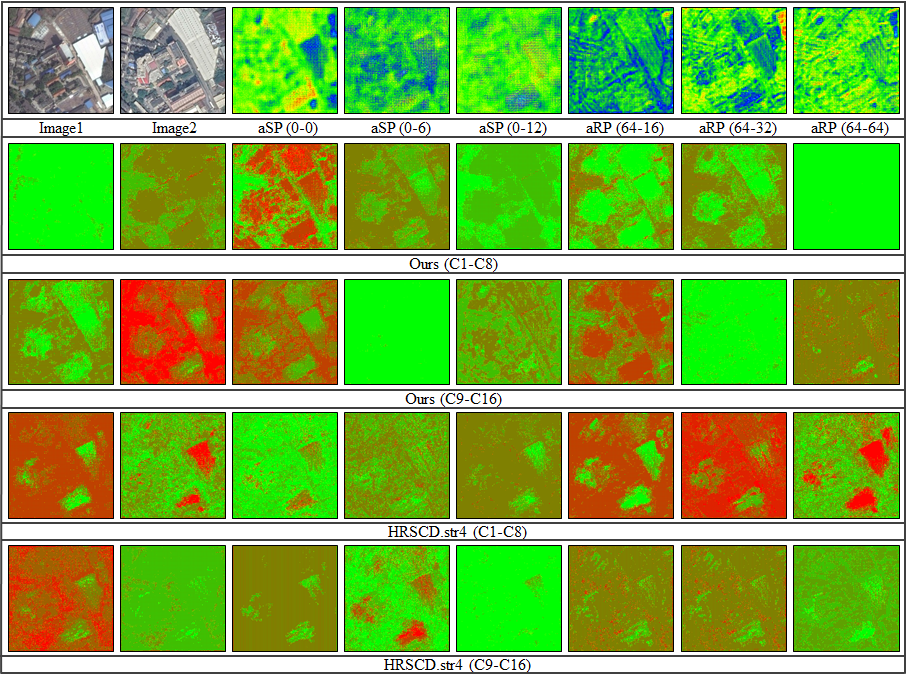}}
		\vspace{-3mm}
		\caption{Feature visualizing in ASN and HRSCD.str4 with residual blocks. The first row shows the mean value of normalized feature samples \emph{w.r.t.} different integrations of asymmetric feature pairs. Features deriving from some asymmetric feature pairs, such as aRP(64-32) in the top right, contain more distinguishable values in regions of asymmetric changes. The rest rows show all 16 channels of corresponding features in the change detection branch of ASN and HRSCD.str4. C1-C8 means first 8 channels, while C9-C16 means last 8 channels. Similarly, features in ASN contain more distinguishable values in regions of asymmetric changes compared with HRSCD.str4.} 
		\label{fig:Feature}
	\end{figure*}
	\begin{figure*}[!t]
		\centering
		{\includegraphics[width=0.74\linewidth]{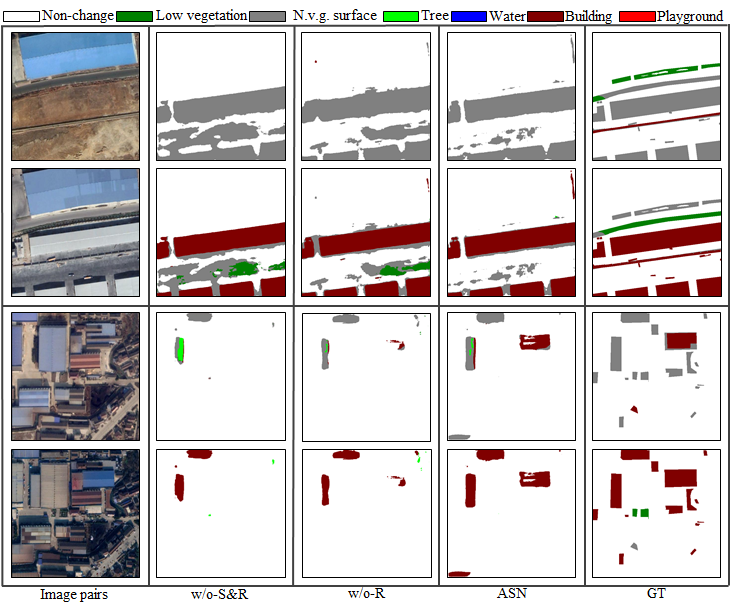}}	
		\vspace{-3mm}
		\caption{Results of ablation study with residual blocks. With our proposed modules, the model can better extract illegible change types and alleviate the false identifications of changed pixels.}
		\label{fig:ablation}
	\end{figure*}
	
	To further explore the detail comparison, we list the categorical SeK (calculated from $2 \times 2$ confusion matrix concentrating on each change type) and categorical intersection over union ($\textrm{IOU}_1$ and $\textrm{IOU}_2$) in Tab.\ref{Tab-Detail-Residual}. Each row (except \emph{non-change}) indicates land-cover class in the first image, while each column indicates land-cover class in the second image. Categorical SeK evaluates identification of all change types except \emph{non-change}, while categorical intersection over union measures the results of \emph{non-change} pixel extraction. We choose HRSCD.str4, HRSCD.str1 and ASN-ATL to be listed in Tab.~\ref{Tab-Detail-Residual} for detail comparsion. We can see that ASN-ATL achieves the best performance on an overwhelming majority of change types under all kinds of testing strategies. 
	
	\begin{table}[htb!]
		\scriptsize
		\centering
		\renewcommand\tabcolsep{3.9pt} 
		\small
		\caption{Ablation study on asymmetric feature pairs with residual backbone. ASN-w/o-AF represents retaining multi-scale structures without the asymmetric feature pair integration.}
		\vspace{-2mm}
		\label{tab:af}
		\setlength{\arraycolsep}{0.1pt}
		\begin{threeparttable}
			\begin{tabular}{c|cc|cc|cc}
				
				\toprule
				
				\multirow{2}{*}{Methods} & \multicolumn{2}{c}{MS \underline{\space\space\space} Flip \underline{\space\space\space}} & \multicolumn{2}{c}{MS \underline{\checkmark} Flip \underline{\space\space\space}} & \multicolumn{2}{c}{MS \underline{\checkmark} Flip \underline{\checkmark} } \\
				
				\cmidrule(r){2-3} \cmidrule(r){4-5} \cmidrule(r){6-7}
				
				&mIOU&SeK&mIOU&SeK&mIOU&SeK\\
				
				\midrule
				
				ASN-w/o-AF& 67.7& 13.5& 68.2& 14.4&68.3&14.5\\	
				ASN&\textbf{69.0}&\textbf{15.2}&\textbf{69.5}&\textbf{16.1}&\textbf{69.7}&\textbf{16.2}\\	
				\bottomrule
			\end{tabular}
		\end{threeparttable}
	\end{table}
	
	\begin{table*}[htb!]
		\scriptsize
		\centering
		\renewcommand\tabcolsep{3.9pt} 
		\footnotesize
		\caption{Ablation study of ASN. w/o-R\&S represents without aSP and aRP, while w/o-R represents without aRP.}
		\vspace{-3mm}
		\label{tab:Ablation}
		\setlength{\arraycolsep}{0.1pt}
		\begin{threeparttable}
			\begin{tabular}{c|ccc|ccc|ccc}
				
				\toprule
				
				\multirow{2}{*}{Ablation} & \multicolumn{3}{c}{Residual} & \multicolumn{3}{c}{Xception} & \multicolumn{3}{c}{SENet} \\
				
				\cmidrule(r){2-4} \cmidrule(r){5-7} \cmidrule(r){8-10}
				
				study&w/o-R\&S&w/o-R&ASN&w/o-R\&S&w/o-R&ASN&w/o-R\&S&w/o-R&ASN\\
				
				\midrule
				
				mIOU&68.7& 68.9 &\textbf{69.7}&68.3& 68.8 &\textbf{69.0}&68.5& 69.2 &\textbf{69.7}\\
				
				SeK&14.8& 15.4  &\textbf{16.2}&14.6& 15.4 &\textbf{15.5}&15.0& 15.8 &\textbf{16.6}\\

				\bottomrule
			\end{tabular}
		\end{threeparttable}
	\end{table*}
	
	We then substitute all the basic residual blocks in the encoder with Xception blocks and Squeeze-and-Excitation blocks. As illustrated in Tab.\ref{tab:Overall-Comparison-Xception} and Tab.\ref{tab:Overall-Comparison-SENet}, our proposed ASN still outperforms other models under all kinds of testing processes. Especially, under full testing strategies, ASN-ATL outperforms HRSCD.str4 by 1.9 in mIOU and 2.2 in SeK with the Xception blocks, while HRSCD.str4 is inferior to ASN-ATL by 2.0 in mIOU and 1.9 in SeK with Squeeze-and-Excitation blocks. Meanwhile, ATL provides 0.9 improvements in mIOU and 0.8 improvements in SeK with Xception blocks, while 0.5 improvements in mIOU and 0.7 improvements in SeK can be seen with Squeeze-and-Excitation blocks. We can conclude that ASN keeps the outperformances with all kinds of backbones and ATL does improve the model performances.
	Moreover, Fig.\ref{fig:Residual}-\ref{fig:SENet} report the visual results of HRSCD.str4 and ASN with all three utilized basic blocks. We list the results of the same data samples in the left parts of these figures, which shows the stable effectiveness of ASN with different encoder backbones. Also, we list the results of different data samples in the right parts to exhibit the generalization abilities of the proposed modules.
	
	
	\subsection{Visualization of Learned Features}\label{FV}	
	Recalling the discussion in Sec.\ref{sec:introduction}, we can see in Fig.\ref{fig:echo} that ASN alleviates the categorical ambiguity caused by asymmetric changes. Moreover, as illustrated in Fig.\ref{fig:Feature}, samples of feature maps in ASN are more likely to contain distinguishable response values in changed regions compared with HRSCD.str4. 
	Specifically, the first row shows some mean value maps of normalized features integrated by different feature pairs. Among these samples, we can see that some feature maps in aSP integrated by asymmetric feature pairs (0-6 and 0-12) contain more distinguishable response values in asymmetric changes (framed out in Fig.\ref{fig:echo}).
	Similarly, feature map calculated from asymmetric feature pairs in aRP (64-32) also contains more distinguishable values in changed regions.
	Moreover, we can conclude from the rest rows of Fig.\ref{fig:Feature} that feature maps integrated by asymmetric feature pairs in change detection branch are also more likely to contain different response values in asymmetric changes, while corresponding features in HRSCD.str4 contain similar response values in asymmetric changes with other unchanged regions.

	
	
	\subsection{Ablation Study}\label{Ablation}	
	In order to further explore the validity of the proposed modules, we remove aRP and aSP successively to set up the ablation study and check corresponding influences on model performance. As illustrated in Tab.\ref{tab:Ablation}, aSP leads to 0.2 improvements in mIOU and 0.6 improvements in SeK with the residual blocks. Then, aRP further improves the model performance by 0.8 in mIOU and SeK. Besides, with the Xception block, aSP improves mIOU by 0.5 and SeK by 0.8, while aRP further improves mIOU by 0.2 and SeK by 0.1. Similarly, with the Squeeze-and-Excitation block, aSP brings about 0.7 improvements in mIOU and 0.8 improvements in SeK, while aRP raises 0.5 improvements in mIOU and 0.8 improvements in SeK. In summary, aSP and aRP improve model performances on BCD and SCD simultaneously. The effectiveness of our proposed modules can be verified.
	
	Moreover, Fig.\ref{fig:ablation} shows the visual results of the ablation study, where we can see that our proposed modules can better extract illegible changed regions, such as the changes between {\emph{buildings}} across multi-temporal images. In the meantime, the false positive identification of changed regions could be reduced with the proposed modules.
	
	Further to explore the effects of asymmetric feature pairs, we remove feature pair integrations and retain the multi-scale structures in aSP to see the influence on results with residual backbones. As we can see in Tab.\ref{tab:af}, the incomplete model suffers a decline in the performance, which verifies merits of the integration of asymmetric feature pairs.

	\section{Conclusion}
	In this paper, we propose an asymmetric siamese network for semantic change detection to alleviate categorical ambiguity caused by asymmetric changes through locally asymmetric architecture. To better train deep models, we create a large scale well-annotated SECOND as a new benchmark, which includes the changed regions between the same land-cover class. Further, to alleviate the influence of label imbalance during model training and evaluation, we design an adaptive threshold learning module and an SeK to adaptively revise the output deflections and fix unreasonable scores computed with traditional metrics respectively. The experimental results show that the proposed model stably achieves the best results with different encoder backbones compared with state-of-the-art algorithms.

	
	%

	\ifCLASSOPTIONcaptionsoff
	\newpage
	\fi

	
	
	%
	{\small
		\bibliographystyle{IEEEtran}
		\bibliography{egbib}
	}
	
	%
	
	\ifCLASSOPTIONcaptionsoff
	\newpage
	\vskip -1.5\baselineskip plus -1fil
	
	\begin{IEEEbiography}[{\includegraphics[width=1in,height=1.25in,clip,keepaspectratio]{images/ykp.png}}]{Kunping Yang} received his B.S. degree and M.S. degree in Mathematics from Wuhan University, Wuhan, China, in 2014 and 2016 respectively. He is currently pursuing his Ph.D. degree in the State Key Laboratory of Information Engineering in Surveying, Mapping and Remote Sensing (LIESMARS) at Wuhan University. His research interests include mathematical modeling of images, remote sensing image understanding, semantic segmentation and change detection.
	\end{IEEEbiography}
	
	\vskip -1.9\baselineskip plus -1fil
	
	\begin{IEEEbiography}[{\includegraphics[width=1in,height=1.25in,clip,keepaspectratio]{images/xgs.jpg}}]{Gui-Song Xia}
		(M'10-SM'15) received his Ph.D. degree in image processing and computer vision from CNRS LTCI, T{\'e}l{\'e}com ParisTech, Paris, France, in 2011. From 2011 to 2012, he has been a Post-Doctoral Researcher with the Centre de Recherche en Math{\'e}matiques de la Decision, CNRS, Paris-Dauphine University, Paris, for one and a half years. He has also been working as Visiting Scholar at DMA, {\'E}cole Normale Sup{\'e}rieure (ENS-Paris) in 2018.
		
		He is currently working as a full professor in computer vision and photogrammetry jointly in the School of Computer Science and the State Key Lab. of LIESMARS at Wuhan University, where he is leading a research team named Computational and Photogrammetric Vision Team (CAPTAIN), working toward developing mathematical and computational models to measure and understand our physic environments with vision information. His current research interests include mathematical modeling of images and videos, structure from motion, perceptual grouping, and remote sensing image interpretation. He serves on the  Editorial Boards of the journals Pattern Recognition, Signal Processing: Image Communications, EURASIP Journal on Image \& Video Processing, and Journal of Remote Sensing. He has also served as Guest Editors for journals including IEEE Trans. on Big Data, Pattern Recognition Letter, {\em etc.}.
	\end{IEEEbiography}
	
	\vskip -1.9\baselineskip plus -1fil
	
	\begin{IEEEbiography}[{\includegraphics[width=1in,height=1.25in,clip,keepaspectratio]{images/lzc.png}}]{Zicheng Liu} received his B.S. degree in Electrical Engineering from Wuhan University of Science and Technology, Wuhan, China, in 2018. He is currently pursuing his M.S. degree in the State Key Laboratory of Information Engineering in Surveying, Mapping and Remote Sensing (LIESMARS) at Wuhan University. His research interests include remote sensing image understanding and change detection.
	\end{IEEEbiography}
	
	\vskip -1.9\baselineskip plus -1fil
	
	\begin{IEEEbiography}[{\includegraphics[width=1in,height=1.25in,clip,keepaspectratio]{images/dubo.jpg}}]{Bo Du}
		(M’10–SM’15) received the Ph.D. degree  from the State Key Lab. LIESMARS, Wuhan University, Wuhan, China, in 2010.
		He is currently a professor with the National Engineering Research Center for Multimedia Software, Institute of Artificial Intelligence and School of Computer Science, Wuhan University, China. He has more than 60 research papers published in the IEEE Trans. on Neural Networks and Learning System (TNNLS), IEEE Trans. on image processing (TIP), and IEEE Trans. on Multimedia (TMM), IEEE Trans. on Geoscience and Remote Sensing (TGRS), IEEE Journal of Selected Topics in Earth Observations and Applied Remote Sensing (JSTARS), and IEEE Geoscience and Remote Sensing Letters (GRSL), etc. Thirteen of them are ESI hot papers or highly cited ones. His major research interests include pattern recognition, hyperspectral image processing, machine learning and signal processing.
		
		He is currently a senior member of IEEE. He received 2020 Best Paper award for IEEE TGRS, Highly Cited Researcher 2019 award from Web of Science Group, the distinguished paper award from IJCAI 2018, the best paper award of IEEE Whispers 2018 and the champion award of the IEEE Data Fusion Contest 2018. He received ACM Rising Star Awards Wuhan in 2015. He serves as associate editor for Pattern Recognition and Neurocomputing, senior PC/PC for IJCAI/AAAI/KDD, and as area chair for ICPR and IJCNN. He was the Session Chair for both International Geoscience and Remote Sensing Symposium (IGARSS) 2018/2016 and the 4th IEEE GRSS Workshop on Hyperspectral Image and Signal Processing: Evolution in Remote Sensing. 
	\end{IEEEbiography}
	
	\vskip -1.9\baselineskip plus -1fil
	
	\begin{IEEEbiography}[{\includegraphics[width=1in,height=1.25in,clip,keepaspectratio]{images/wenyang.jpg}}]{Wen Yang}
		(M'09-SM'16) received a B.S. degree in electronic apparatus and surveying technology, a M.S. degree in computer application technology and a Ph.D. degree in communication and information system from Wuhan University, Wuhan, China, in 1998, 2001, and 2004, respectively. From 2008 to 2009, he worked as a Visiting Scholar with the Apprentissage et Interfaces (AI) Team, Laboratoire Jean Kuntzmann (LJK), Grenoble, France. From 2010 to 2013, he worked as a Post-doctoral Researcher with the State Key Lab. LIESMARS, Wuhan University. Since then, he has been a Full Professor with the School of Electronic Information, Wuhan University. His research interests include object detection and recognition, semantic segmentation and multisensor information fusion.
	\end{IEEEbiography}
	
	\vskip -1.9\baselineskip plus -1fil
	
	\begin{IEEEbiography}[{\includegraphics[width=1in,height=1.25in,clip,keepaspectratio]{images/Marcello.png}}]{Marcello Pelillo} (SM'04-F'13) 
		is a Full Professor of Computer Science at Ca’ Foscari University,
		Venice, where he leads the Computer Vision and Pattern Recognition Lab. He has been the Director of the European Centre for Living
		Technology (ECLT) and has held visiting research/teaching positions in several
		institutions including Yale University (USA), University College London (UK), McGill
		University (Canada), University of Vienna (Austria), York University (UK), NICTA
		(Australia), Wuhan University (China), Huazhong University of Science and Technology
		(China), and South China University of Technology (China). He is also
		an external affiliate of the Computer Science Department at Drexel University (USA). 
		His research interests are in the areas of computer vision, machine learning and pattern recognition where he has published more than 200 technical papers in refereed journals, handbooks, and conference proceedings.
		
		He has been General Chair for ICCV 2017, Program Chair for ICPR 2020, and has been Track or Area Chair for several conferences in his area. 
		He is the Specialty Chief Editor of Frontiers in Computer Vision and serves, or has served, on the Editorial Boards of several journals, including IEEE Transactions on Pattern Analysis and Machine Intelligence, Pattern Recognition, IET Computer Vision, and Brain Informatics.
		He also serves on the Advisory Board of Springer’s International Journal of Machine Learning and Cybernetics.
		Prof. Pelillo has been elected Fellow of the IEEE and Fellow of the IAPR and is an IEEE SMC Distinguished Lecturer. His Erd\"os number is 2.
		
	\end{IEEEbiography}
	
	
	
	
	\fi
	
\end{document}